\documentclass[runningheads]{llncs}
\usepackage{graphicx}
\usepackage{comment}
\usepackage{amsmath,amssymb} 
\usepackage{color}
\usepackage{xcolor}
\usepackage{multirow}
\usepackage{diagbox}
\usepackage{wrapfig}
\usepackage{algorithm}
\usepackage{algorithmic}
\usepackage{siunitx}

\begin{document}
\pagestyle{headings}
\mainmatter
\def\ECCVSubNumber{6748}  

\title{Inherent Adversarial Robustness of Deep Spiking Neural Networks: Effects of Discrete Input Encoding and Non-Linear Activations} 

\titlerunning{Adversarial robustness of SNN}
%
\author{Saima Sharmin\inst{1}\orcidID{0000-0002-1866-9138} \and
Nitin Rathi\inst{1}\orcidID{0000-0003-0597-064X} \and
Priyadarshini Panda\inst{2}\orcidID{0000-0002-4167-6782} \and
Kaushik Roy\inst{1}\orcidID{0000-0002-0735-9695}}

\authorrunning{S. Sharmin et. al.}
%
\institute{Purdue University, West Lafayette IN 47907, USA\\
\email{\{ssharmin,rathi2,kaushik\}@purdue.edu}\and
Yale University, New Haven CT 06520, USA\\
\email{priya.panda@yale.edu}}

\maketitle
\begin{abstract}
In the recent quest for trustworthy neural networks, we present Spiking Neural Network
(SNN) as a potential candidate for inherent robustness against adversarial attacks.  In this work, we demonstrate that adversarial accuracy of SNNs under gradient-based attacks is higher than their non-spiking counterparts for CIFAR datasets on deep VGG and ResNet architectures, particularly in blackbox attack scenario. We attribute this robustness to two fundamental characteristics of SNNs and analyze their effects. First, we exhibit that input discretization introduced by the Poisson encoder improves adversarial robustness with reduced number of timesteps. Second, we quantify the amount of  adversarial accuracy with increased leak rate in Leaky-Integrate-Fire (LIF) neurons. Our results suggest that SNNs trained with LIF  neurons and smaller number of timesteps are more
robust than the ones with IF (Integrate-Fire) neurons and larger number of timesteps. Also we overcome the bottleneck of creating gradient-based adversarial inputs in temporal domain by proposing a technique for crafting attacks from SNN\footnote[1]{https://github.com/ssharmin/spikingNN-adversarial-attack}.
\end{abstract}
\renewcommand{\labelitemi}{$\bullet$}
\renewcommand{\labelitemii}{$\circ$}
\section{Introduction}
\label{sec:intro}
Adversarial attack is one of the biggest challenges against the success of today's deep neural networks in mission critical applications \cite{kurakin}, \cite{stop}, \cite{malware}. The underlying
concept of an adversarial attack is to purposefully modulate the
input to a neural network such that it is subtle enough to remain
undetectable to human eyes, yet capable of fooling the network
into incorrect decisions. This malicious behavior was first
demonstrated in 2013 by Szegedy {\it{et. al.}} \cite{szegedy} and Biggio {\it{et. al.}} \cite{biggio} in the field of
computer vision  and malware detection, respectively.
Since then, numerous defense mechanisms have been proposed to address this issue. One category of defense includes fine-tuning
the network parameters like adversarial
training \cite{adv_train_goodfellow}, \cite{adv_train_madry},
network distillation \cite{papernot}, stochastic activation pruning \cite{sap} etc. Another category focuses on preprocessing the input before passing through the network like thermometer encoding \cite{thermo_enc}, input quantization \cite{quant_inp}, \cite{priya_quant}, compression \cite{inp_compression} etc. Unfortunately, most of these defense mechanisms have been proved futile by many counter-attack techniques. For example, an ensemble of defenses based on ``gradient-masking" collapsed under the attack proposed in \cite{obfus_grad}. Defensive distillation was broken by Carlini-Wagner method \cite{carlini_dist}, \cite{carlini2}. Adversarial training has the tendency to overfit to the training samples and remain vulnerable to transfer attacks \cite{tramer_ensemble_train}. Hence, the threat of adversarial attack continues to persist.\\ 
In the absence of adversarial robustness in the existing state-of-the-art networks, we feel there is a need for a network with {\it{inherent}} susceptibility against adversarial attacks. In this work, we present Spiking Neural Network (SNN) as a potential candidate due to two of its fundamental distinctions from the non-spiking networks: 
\begin{enumerate}
    \item SNNs operate based on discrete binary data ($0/1$), whereas their non-spiking counterparts, referred as Analog Neural Network (ANN), take in continuous-valued analog signals. Since SNN is a binary spike-based model, input discretization is a constituent element of the network, most commonly done by Poisson encoding.
    \item SNNs employ nonlinear activation function of the biologically inspired Integrate-Fire (IF) or Leaky-Integrate-Fire (LIF) neurons, in contrast to the piecewise-linear ReLU activations used in ANNs.
\end{enumerate}
Among the handful of works done in the field of SNN adversarial attacks \cite{snn_deep_belief}, \cite{snn_adv_train}, most of them are restricted to either simple datasets (MNIST) or shallow networks. However, this work extends to complex datasets (CIFAR) as well as deep SNNs which can achieve comparable accuracy to the state-of-the-art ANNs \cite{abhronil}, \cite{hybrid_nitin}. For robustness comparison with non-spiking networks, we analyze two different types of spiking networks: (1) converted SNN (trained by ANN-SNN conversion \cite{abhronil}) and (2) backpropagated SNN (an ANN-SNN converted network, further incrementally trained by surrogate gradient backpropagation \cite{hybrid_nitin}). We identify that converted SNNs fail to demonstrate more robustness than ANNs. Although authors in \cite{saima_ijcnn} show similar analysis, we explain with experiments the reason behind this discrepancy and, thereby, establish the necessary criteria for an SNN to become adversarially robust. Moreover, we propose an SNN-crafted attack generation technique with the help of the surrogate gradient method. We summarize our contribution as follows:
\begin{itemize}
    \item We show that the adversarial accuracies of SNNs are higher than ANNs under a gradient-based blackbox attack scenario, where the respective clean accuracies are comparable to each other. The attacks were performed on deep VGG and ResNet architectures trained on CIFAR10 and CIFAR100 datasets. For whitebox attacks, the comparison is dependent on the relative strengths of the adversary.
    \item The increased robustness of SNN is attributed to two fundamental characteristics: input discretization through Poisson encoding and non-linear activations of LIF (or IF) neurons.
    \begin{itemize}
    \item We investigate how adversarial accuracy changes with the number of timesteps (inference latency) used in SNN for different levels of input pre-quantization\footnote[2]{Full-precision analog inputs are quantized to lower bit precision values before undergoing the discretization process by the Poisson encoder of SNN}. In case of backpropagated SNNs (trained with smaller number of timesteps), the amount of discretization as well as adversarial robustness increases as we reduce the number of timesteps. Pre-quantization of the analog input brings about further improvement. However, converted SNNs appear to depend only on the input pre-quantization, but invariant to the variation in the number of timesteps. Since these converted SNNs operate under larger number of timesteps \cite{abhronil}, discretization effect is minimized by input averaging, and hence, the observed invariance.
    \item We show that piecewise-linear activation (ReLU) in ANN linearly propagates the adversarial perturbation throughout the network, whereas LIF (or IF) neurons diminish the effect of perturbation at every layer. Additionally, the leak factor in LIF neurons offers an extra knob to control the adversarial perturbation. We perform a quantitative analysis to demonstrate the effect of leak on the adversarial robustness of SNNs.
    \end{itemize}
    Overall, we show that SNNs employing LIF neurons, trained with surrogate gradient-based backpropagation, and operating at less number of timesteps are more robust than SNNs trained with ANN-SNN conversion that requires IF neurons and more number of timesteps. Hence, the training technique plays a crucial role in fulfilling the prerequisites for an adversarially robust SNN.
    \item Gradient-based attack generation in SNNs is non-trivial due to the discontinuous gradient of the LIF (or IF) neurons. We propose a methodology to generate attacks based on the approximate surrogate gradients.
\end{itemize}
\section{Background}
\subsection{Adversarial Attack}
\label{sec:adv}
Given a clean image $x$ belonging to class $i$ and a trained neural network $M$, an adversarial image $x_{adv}$ needs to meet two criteria:
\begin{enumerate}
    \item $x_{adv}$ is visually ``similar" to $x$ {\it{i. e.}} $|x-x_{adv}|=\epsilon$, where $\epsilon$ is a small number.
    \item $x_{adv}$ is misclassified by the neural network, {\it{i. e.}} $M(x_{adv}) \neq i$
\end{enumerate}
The choice of the distance metric $|.|$ depends on the method used to create $x_{adv}$ and $\epsilon$ is a hyper-parameter. In most methods, $l_2$ or $l_{\infty}$ norm is used to measure the similarity and the value of $\epsilon$ is limited to $\leq\frac{8}{255}$ where normalized pixel intensity $x\in [0,1]$ and original $x\in[0,255]$. \\
In this work, we construct adversarial examples using the following two methods:
\subsubsection{Fast Gradient Sign Method (FGSM)} This is one of the simplest methods for constructing adversarial examples, introduced in \cite{adv_train_goodfellow}. For a given instance $x$, true label $y_{true}$ and the corresponding cost function of the network $J(x,y_{true})$, this method aims to search for a perturbation $\delta$ such that it maximizes the cost function for the perturbed input $J(x+\delta,y_{true})$, subject to the constraint $|\delta|_{\infty}<\epsilon$. In closed form, the attack is formulated as,
\begin{equation}
    x_{adv} = x + \epsilon \times sign\big(\nabla_xJ\left(x,y_{true}\right)\big)
    \label{eq:fgsm}
\end{equation}
Here, $\epsilon$ denotes the strength of the attack. 
\subsubsection{Projected Gradient Descent (PGD)} 
This method, proposed in \cite{adv_train_madry}, produces more powerful adversary. PGD is basically a $k\mbox{-}step$ variant of FGSM computed as,
\begin{equation}
    x^{(k+1)}_{adv} =  \Pi_{x+\epsilon}\bigg\{\bigg(x^{(k)}_{adv}+\alpha \times  sign\Big(\nabla_x\big(J(x^{(k)}_{adv},y_{true})\big)\Big)\bigg)\bigg\} 
    \label{eq:pgd}
\end{equation}
where $x^{(0)}_{adv}=x$ and $\alpha (\leq\epsilon)$ refers to the amount of perturbation used per iteration or step, $k$ is the total number of iterations. $\Pi_{x+\epsilon}\{.\}$ performs a projection of its operand on an $\epsilon$-ball around $x$, {\it{i. e.}}, the operand is clipped between $x+\epsilon$ and $x-\epsilon$. Another variant of this method adds a random perturbation of strength $\epsilon$ to $x$ before performing PGD operation. \\ 
 \subsection{Spiking Neural Network (SNN)}
 \label{sec:snn}
 The main difference between an SNN and an ANN is the concept of time. The incoming signals as well as the intermediate node inputs/outputs in an ANN are static analog values, whereas in an SNN, they are binary spikes with a value of $0$ or $1$, which are also functions of time. In the input layer, a Poisson event generation process is used to convert the continuous valued analog signals into binary spike train. Suppose the input image is a 3-D matrix of dimension $h\times w\times l$ with pixel intensity in the range $[0,1]$. At every time step of the SNN operation, a random number (from the normal distribution $\mathcal{N}(0,1)$) is generated for each of these pixels. A spike is triggered at that particular time step if the corresponding pixel intensity is greater than the generated random number. This process continues for a total of $T$ timesteps to produce a spike train for each pixel. Hence, the size of the input spike train is $T \times h \times w \times l$. For a large enough $T$, the timed average of the spike train will be proportional to its analog value. \\
 Every node of the SNN is accompanied with a neuron. Among many neuron models, the most commonly used ones are Integrate-Fire (IF) or Leaky-Integrate-Fire (LIF) neurons. The dynamics of the neuron membrane potential at time $t+1$ is described by,
 \begin{equation}
     V(t+1) = \lambda V(t) + \sum_{i}w_{i}x_{i}(t)
    \label{eq:vt_lif}
 \end{equation}
 Here $\lambda=1$ for IF neurons and $<1$ for LIF neurons. $w_i$ denotes the synaptic weight between current neuron and $i\mbox{-}th$ pre-neuron. $x_i(t)$ is the input spike from the $i\mbox{-}th$ pre-neuron at time $t$. When $V(t+1)$ reaches the threshold voltage $V_{th}$, an output spike is generated and the membrane potential is reset to 0, or in case of soft reset, reduced by the amount of the threshold voltage. At the output layer, inference is performed based on the cumulative membrane potential of the output neurons after the total time $T$ has elapsed. \\
One of the main shortcomings of SNNs is that they are difficult to train, especially for deeper networks. Since the neurons in an SNN have discontinuous gradients, standard gradient-descent techniques do not directly apply. In this work, we use two of the supervised training algorithms \cite{abhronil}, \cite{hybrid_nitin}, which achieve ANN-like accuracy even for deep neural networks and on complex datasets.  
\subsubsection{ANN-SNN Conversion}
\label{sec:conv_snn}
This training algorithm was originally outlined in \cite{diehl} and subsequently improved in \cite{abhronil} for deep networks. Note that the algorithm is suited for training SNNs with IF neurons only. They propose a threshold-balancing technique that sequentially tunes the threshold voltage of each layer. Given a  trained ANN, the ﬁrst step is to generate the input Poisson spike train for the network over the training set for a large enough time-window so that the timed average accurately describes the input. Next, the maximum value of $\sum_iw_ix_i$ (term 2 in Eq.~\ref{eq:vt_lif}) received by layer 1 is recorded over the entire time range for several minibatches of the training data. This is referred as the maximum activation for layer 1. The threshold value of layer 1 neuron is replaced by this maximum activation keeping the synaptic weights unchanged. Such threshold tuning operation ensures that the IF neuron activity precisely mimics the ReLU function in the corresponding ANN. After balancing layer 1 threshold, the method is continued for all subsequent layers sequentially.

\subsubsection{Conversion $\&$ Surrogate-Gradient Backpropagation}
\label{sec:hybrid_snn}
\begin{wrapfigure}{r}{0.32\textwidth}
  \begin{center}
    \includegraphics[width=0.3\textwidth]{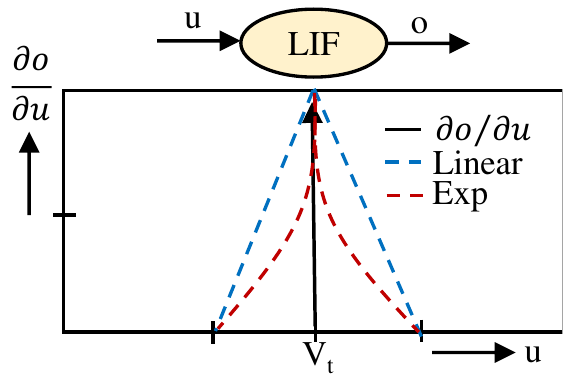}
  \end{center}
  \caption{Surrogate gradient approximation of an LIF neuron.}
  \label{fig:surr_grad}
\end{wrapfigure}
In order to take advantage of the standard backpropagation-based optimization procedures, authors in \cite{surr_grad}, \cite{surr_grad1}, \cite{surr_grad2} introduced the surrogate gradient technique. The input-output characteristics of an LIF (or IF) neuron is a step function, the gradient of which is discontinuous at the threshold point (Fig. \ref{fig:surr_grad}). In surrogate gradient technique, the gradient is approximated by pseudo-derivatives like linear or exponential functions. Authors in \cite{hybrid_nitin} proposed a novel approximation function for these gradients by utilizing the spike time information in the derivative. The gradient at timestep $t$ is computed as follows:
\begin{equation}
    \frac{\partial o^t}{\partial u^t} = \alpha e^{-\beta \Delta t}
    \label{eq:surr_grad}
\end{equation}
Here, $o^t$ is the output spike at time $t$, $u^t$ is the membrane potential at $t$, $\Delta t$ is the difference between current timestep and the last timestep post-neuron generated a spike. $\alpha$ and $\beta$ are hyperparameters. Once the neuron gradients are approximated, backpropagation through time (BPTT) \cite{bptt} is performed using the chain rule of derivatives. In BPTT, the network is unrolled over all timesteps. The final output is computed as the cumulation of outputs at every timestep and eventually, loss is defined on the summed output. During backward propagation of the loss, the gradients are accumulated over time and used in gradient-descent optimization. Authors in \cite{hybrid_nitin} proposed a hybrid training procedure in which the surrogate-gradient training is preceded by an ANN-SNN conversion to initialize the weights and thresholds of the network. The advantage of this method over ANN-SNN conversion is twofold: one can train a network with both IF and LIF neurons and the number of timesteps required for training is reduced by a factor of 10 without losing accuracy.
\section{Experiments}
\subsection{Dataset and Models}
\label{sec:data_model}
We conduct our experiments on VGG5 and ResNet20 for CIFAR10 dataset and VGG11 for CIFAR100. The network topology for VGG5 consists of conv3,64-avgpool-conv3,128 ($\times$2)-avgpool-fc1024 ($\times$2)-fc10. Here conv3,64 refers to a convolutional layer with 64 output filters and 3$\times$3 kernel size. fc1024 is a fully-connected layer with 1024 output neurons. VGG11 contains 11 weight layers corresponding to the configuration A in \cite{vgg} with maxpool layers replaced by average pooling. ResNet20 follows the proposed architecture for CIFAR10 in \cite{resnet}, except the initial $7\times 7$ non-residual convolutional layer is replaced by a series of two $3\times 3$ convolutional layers. For ANN-SNN conversion of ResNet20, threshold balancing is performed only on these initial non-residual units (as demonstrated by \cite{abhronil}). The neurons (in both ANN and SNN) contain no bias terms, since they have an indirect effect on the computation of threshold voltages during ANN-SNN conversion. The absence of bias eliminates the use of batch normalization \cite{bn} as a regularizer. Instead, a dropout layer is used after every ReLU (except for those which are followed by a pooling layer). 
\subsection{Training Procedure}
\label{sec:training}
The aim of our experiment is to compare adversarial attack on 3 networks: 1) ANN, 2) SNN trained by ANN-SNN conversion and 3) SNN trained by backpropagation, with initial conversion. These networks will be referred as ANN, SNN-conv and SNN-BP, respectively from this point onward. \\
For both CIFAR10 and CIFAR100 datasets, we follow the data augmentation  techniques in \cite{data_aug}: 4 pixels are padded on each side, and a 32$\times$32 crop is randomly sampled from the padded image or its horizontal ﬂip. Testing is performed on the original 32 $\times$ 32 images. Both training and testing data are normalized to $[0,1]$. For training the ANNs, we use cross-entropy loss with stochastic gradient descent optimization (weight decay=0.0001, mometum=0.9). VGG5 (and ResNet20) are trained for a total of 200 epochs, with an initial learning rate of 0.1 (0.05), which is divided by 10 at 100-th (80-th) and 150-th (120-th) epoch. VGG11 with CIFAR100 is trained for 250 epochs with similar learning schedule. 
During training SNN-conv networks, a total of 2500 timesteps are used for all VGG and ResNet architectures. 
SNN-BP networks are trained for 15 epochs with cross-entropy loss and adam \cite{adam} optimizer (weight decay=0.0005). Initial learning rate is 0.0001, which is halved every 5 epochs. A total of 100 timesteps is used for VGG5 and 200 timesteps for ResNet20 \& VGG11. Training is performed with either linear surrogate gradient approximation \cite{surr_grad} or spike time dependent approximation \cite{hybrid_nitin} with $\alpha$ = 0.3, $\beta$=0.01 (in Eq.~\ref{eq:surr_grad}). Both techniques yield approximately similar results. Leak factor $\lambda$ is kept at 0.99 in all cases, except in the analysis for the leak effect.\\
In order to analyze the effect of input quantization (with varying number of timesteps) and leak factors, only VGG5 networks with CIFAR10 dataset is used. 

\subsection{Adversarial Input Generation Methodology}
\label{sec:adv_method}
For the purpose of whitebox attacks, we  need to construct adversarial samples from all three networks (ANN, SNN-conv, SNN-BP). The ANN-crafted FGSM and PGD attacks are generated using the standard techniques described in Eq.~\ref{eq:fgsm} and \ref{eq:pgd}, respectively. We carry out non-targeted attacks with $\epsilon=8/255$. PGD attacks are performed with iteration steps $k=7$ and per-step perturbation $\alpha=2/255$. 
FGSM or PGD method cannot be directly applied to SNN due to its discontinuous gradient problem (described in Sec. \ref{sec:hybrid_snn}). To that end, we outline a surrogate-gradient based FGSM (and PGD) technique. In SNN, analog input $X$ is converted to Poisson spike train $X_{spike}$ which is fed into the $1st$ convolutional layer. If $X_{rate}$ is the timed average of $X_{spike}$, the membrane potential of the $1st$ convolutional layer $X_{conv1}$ can be approximated as
\begin{equation}
    X_{conv1} \approx Conv(X_{rate}, W_{conv1})
\end{equation}
$W_{conv1}$ is the weight of the $1st$ convolutional layer. From this equation, the sign of the gradient of the network loss function $J$ w.r.t. $X_{rate}$ or $X$ is described by (detailed derivation is provided in {\it supplementary}),
\begin{equation}
    sign\Big(\frac{\partial J}{\partial X}\Big) \approx sign\Big(\frac{\partial J}{\partial X_{rate}}\Big) = sign\Big(Conv\big(\frac{\partial J}{\partial X_{conv1}}, W_{conv1}^{180 rotated}\big)\Big)
    \label{eq:snn_crafted_fgsm}
\end{equation}
Surrogate gradient technique yields $\frac{\partial J}{\partial X_{conv1}}$ from SNN, which is plugged into Eq.~\ref{eq:snn_crafted_fgsm} to calculate the sign of the input gradient. This sign matrix is later used to compute $X_{adv}$ according to standard FGSM or PGD method. The algorithm is summarized in \ref{algo:snn_crafted}. 
\begin{algorithm}
\caption{SNN-crafted $X_{adv}: FGSM$}
\label{algo:snn_crafted}
\begin{algorithmic} 
\REQUIRE Input ($X,y_{true}$), Trained SNN ($N$) with loss function J.
\ENSURE $\frac{\partial J}{\partial X_{conv1}} \leftarrow 0$
\FOR{timestep $t$ in total time $T$}
\STATE{\textbf{forward:} Loss $J(X,y_{true})$}
\STATE{\textbf{backward :} Accumulate gradient $\frac{\partial J}{\partial X_{conv1}} += X_{conv1}.grad$}
\ENDFOR
\STATE{\textbf{post-processing:} $sign(\frac{\partial J}{\partial X})= sign\Big(Conv(\frac{\partial J}{\partial X_{conv1}}, W_{conv1}^{180rotated}$)}\Big)
\STATE{\textbf{SNN-crafted adversary:} $X_{adv}^{SNN} = X + \epsilon \times sign(\frac{\partial J}{\partial X})$}
\end{algorithmic}
\end{algorithm}
\section{Results}
\subsection{ANN vs SNN}
\label{sec:annVSsnn}
Table~\ref{table:annVSsnn} summarizes our results for CIFAR10 (VGG5 \& ResNet20) and CIFAR100 (VGG11) datasets in whitebox and blackbox settings. For each architecture, we start with three networks: ANN, SNN-conv and SNN-BP, trained to achieve comparable baseline clean accuracy. Let us refer them as $M_{ANN}$, $M_{SNN\mbox{-}conv}$ and $M_{SNN\mbox{-}BP}$, respectively. During blackbox attack, we generate an adversarial test dataset $x_{adv}$ from a separately trained ANN of the same network topology as the target model but different initialization. It is clear that the adversarial accuracy of SNN-BP during FGSM and PGD blackbox attacks is higher than the corresponding ANN and SNN-conv models, irrespective of the size of the dataset or network architecture (the highest value of the accuracy for each attack case is highlighted by {\it orange text}). The amount of improvement in adversarial accuracy, compared to ANN, is listed as $\Delta$ in the Table. If $M_{ANN}$ and $M_{SNN\mbox{-}BP}$ yield adversarial accuracy of $p_{ANN}\%$ and $p_{SNN\mbox{-}BP}\%$, respectively, the value of $\Delta$ amounts to $p_{SNN\mbox{-}BP}\%-p_{ANN}\%$. On the other hand, during whitebox attack, we generate three sets of adversarial test dataset: $x_{adv,ANN}$ (generated from $M_{ANN}$), $x_{adv,SNN\mbox{-}conv}$ (generated from $M_{SNN\mbox{-}conv}$) and so on. Since ANN and SNN  have widely different operating dynamics and constituent elements, the strength of the constructed adversary varies significantly from ANN to SNN during whitebox attack (demonstrated in Sec. \ref{sec:annVSsnn_attack}). SNN-BP shows significant improvement in whitebox adversarial accuracy ($\Delta$ ranging from 2\% to 4.6\%) for both VGG and Resnet architectures with CIFAR10 dataset. In contrast, VGG11 ANN with CIFAR100 manifests higher whitebox accuracy than SNN-BP. We attribute this discrepancy to the difference in adversary-strength of ANN \& SNN for different dataset and network architectures. 
\renewcommand{\arraystretch}{1.4}
\setlength\arrayrulewidth{0.6pt}
\setlength{\tabcolsep}{4pt}
\begin{table}[t]
\begin{center}
\caption{A comparison of the clean and adversarial (FGSM and PGD) test accuracy (\%) among ANN, SNN-conv and SNN-BP networks. Highest value of the accuracy for each attack case is marked in {\it orange text}. FGSM accuracy is calculated at $\epsilon=8/255$. For PGD, $\epsilon=8/255$, $\alpha$ (per-step perturbation) $=2/255$, $k$ (number of steps) $= 7$. The blackbox attacks are generated from a separately trained ANN of the same network topology as the target model but different initialization }
\label{table:annVSsnn}
\begin{tabular}{|c|c||cp{9mm}p{9mm}|c||cp{9mm}p{9mm}|c|}
\hline
  &  & \multicolumn{4}{c||}{Whitebox} & \multicolumn{4}{c|}{Blackbox} \\
\hline
  &  & ANN & SNN- \newline conv & SNN- \newline BP & $\Delta^\dagger$ & ANN & SNN-\newline conv & SNN- \newline BP& $\Delta^\dagger$ \\
\hline\hline
\multicolumn{9}{c}{CIFAR10} \\
\hline\hline 
\multirow{3}{*}{\rotatebox{90}{VGG5}} & Clean & 90\% & 89.9\% & 89.3\% & $\mbox{---}$ & 90\% & 89.9\% & 89.3\% & $\mbox{---}$ \\ 
& FGSM & 10.4\% & 7.7\% & \textcolor{orange}{15\%} & \textbf{4.6\%} & 18.9\% & 19.3\% & \textcolor{orange}{21.5\%} & \textbf{2.6\%} \\ 
 & PGD & 1.8\% & 1.7\% & \textcolor{orange}{3.8\%} & \textbf{2.0\%} & 9.3\% & 9.6\% & \textcolor{orange}{16.0\%} & \textbf{6.7\%} \\
\hline
\multirow{3}{*}{\rotatebox{90}{ResNet20}} & Clean & 88.0\% & 87.5\% & 86.1\% & $\mbox{---}$ & 88.0\% & 87.5\% & 86.1\% & $\mbox{---}$ \\ 
& FGSM & 28.9\% & 28.8\% & \textcolor{orange}{31.3\%} & \textbf{2.4\%} & 56.7\% & \textcolor{orange}{56.8\%} & \textcolor{orange}{56.8\%} & \textbf{0.1\%} \\ 
 & PGD & 1.9\% & 1.4\% & \textcolor{orange}{4.9\%} & \textbf{3.0\%} & 41.5\% & 41.6\% & \textcolor{orange}{46.5\%} & \textbf{5.0\%} \\
\hline\hline
 \multicolumn{9}{c}{CIFAR100} \\
 \hline\hline
 \multirow{3}{*}{\rotatebox{90}{VGG11}}  & Clean & 67.1\% & 66.8\% & 64.4\% & $\mbox{---}$ & 67.1\% & 66.8\% & 64.4\% & $\mbox{---}$ \\ 
 & FGSM & \textcolor{orange}{17.1\%} & 10.5\% & 15.5\% & \textbf{-1.6\%} & 21.2\% & \textcolor{orange}{21.4\%} & \textcolor{orange}{21.4\%} & \textbf{0.2\%} \\ 
 & PGD & \textcolor{orange}{8.5\%} & 4.1\% & 6.3\% & \textbf{-2.2\%} & 15.6\% & 15.8\% & \textcolor{orange}{16.5\%} & \textbf{0.9\%} \\
\hline
  \multicolumn{9}{l}{$^\dagger\Delta$ = Adversarial accuracy (SNN-BP) - Adversarial accuracy (ANN)}
\end{tabular}
\end{center}
\end{table}
From Table~\ref{table:annVSsnn}, it is evident that SNN-BP networks exhibit the highest amount of adversarial accuracy ({\it{orange text}}) among the three networks in all blackbox attack cases (attacked by a common adversary), whereas SNN-conv and ANN demonstrate comparable accuracy, irrespective of the dataset, network topology or attack generation method. Hence, we conclude that SNN-BP is inherently more robust compared to their non-spiking counterpart as well as SNN-conv models, when all three networks are attacked by identical adversarial inputs. It is important to mention here that our conclusion is validated for VGG \& ResNet architectures and gradient-based attacks only.\\ In the next two subsections, we explain two characteristics of SNNs contributing towards this robustness, as well as the reason for SNN-conv not being able to show similar behavior.
\subsubsection{Effect of input quantization and number of timesteps}
\label{sec:poisson}
The main idea behind non-linear input pre-processing as a defense mechanism is to discretize continuous-valued input signals so that the network becomes non-transparent to adversarial perturbations, as long as they lie within the discretization bin. SNN is a binary spike-based network, which demands encoding any analog valued input signal into binary spike train, and hence, we believe it has the inherent robustness. In our SNN models, we employ Poisson rate encoding, where the output spike rate is proportional to the input pixel intensity. However, the amount of discretization introduced by the Poisson encoder varies with the number of timesteps used. Hence, the adversarial accuracy of the network can be controlled by varying the number of timesteps as long as the clean accuracy remains within reasonable limit. This effect can be further enhanced by quantizing the analog input before feeding into the Poisson encoder. 
\begin{figure}[t]
\centering
\includegraphics[height=4.0cm]{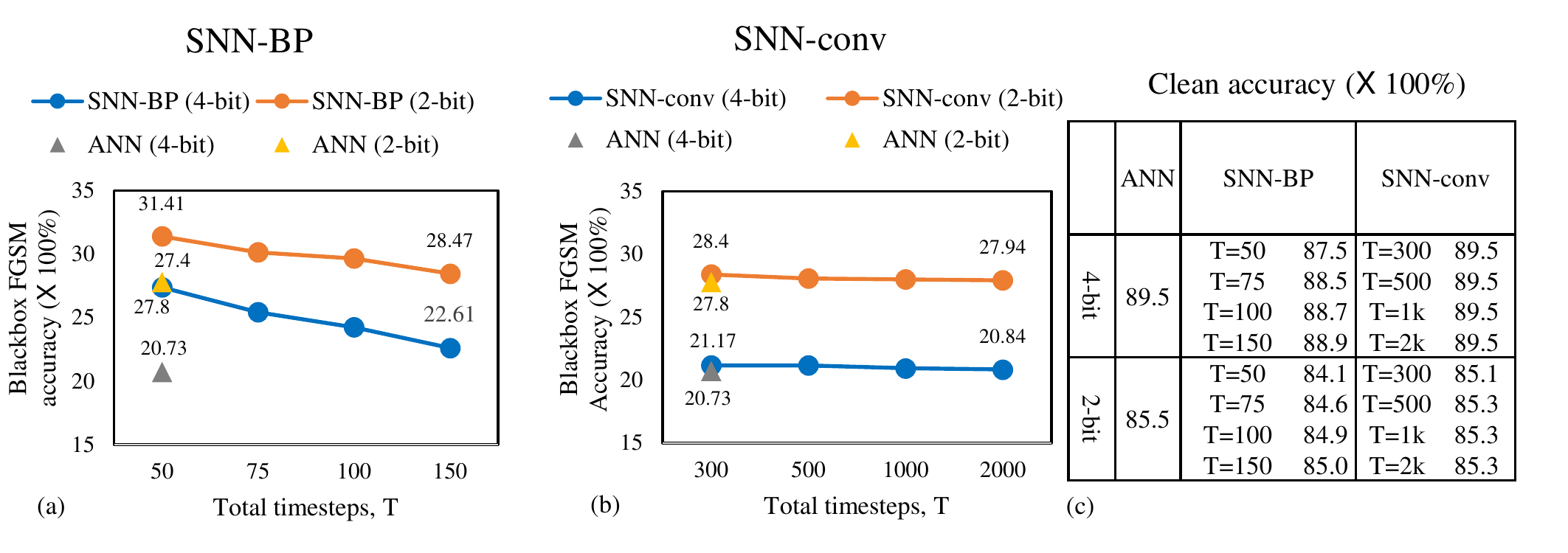}
\caption{Blackbox FGSM accuracy(\%) versus total number of timesteps ($T$) plot with 4-bit ({\it blue}) and 2-bit ({\it orange}) input quantization for (a) SNN-BP and (b) SNN-conv networks. SNN-BP adversarial accuracy increases drastically with decreased number of timesteps, whereas SNN-conv is insensitive to it. (c) A table summarizing the clean accuracy of ANN, SNN-conv and SNN-BP for different input quantizations and number of timesteps.}
\label{fig:poisson_fig2}
\end{figure}
In Fig.~\ref{fig:poisson_fig2}(a), we demonstrate the FGSM adversarial accuracy of an SNN-BP network (VGG5) trained for 50, 75, 100 and 150 timesteps with CIFAR10 dataset. As number of timesteps drop from 150 to 50, accuracy increases by $\sim5$\% ({\it{blue line}}) for 4-bit input quantization. Note that clean accuracy drops by only 1.4\% within this range, from 88.9\% (150 timesteps) to 87.5\% (50 timesteps), as showed in the table in Fig.~\ref{fig:poisson_fig2}(c). Additional reduction of the number of timesteps leads to larger degradation of clean accuracy. The adversarial accuracy for corresponding ANN (with 4-bit input quantization) is showed in {\it{gray triangle}} in the same plot for comparison. Further increase in adversarial accuracy is obtained by pre-quantizing the analog inputs to 2-bits ({\it{orange line}}) and it follows the same trend with number of timesteps. Thus varying the number of timesteps introduces an extra knob for controlling the level of discretization in SNN-BP in addition to the input pre-quantizations.
In contrast, in Fig.~\ref{fig:poisson_fig2}(b), similar experiments performed on SNN-conv network demonstrate little increase in adversarial accuracy with the number of timesteps. Only pre-quantization of input signal causes improvement of accuracy from $\sim21$\% to $\sim28$\%. Note that the range of the number of timesteps used for SNN-conv (300 to 2000) is much higher than SNN-BP, because converted networks have higher inference latency.
\begin{figure}[t]
\centering
\includegraphics[height=4.0cm]{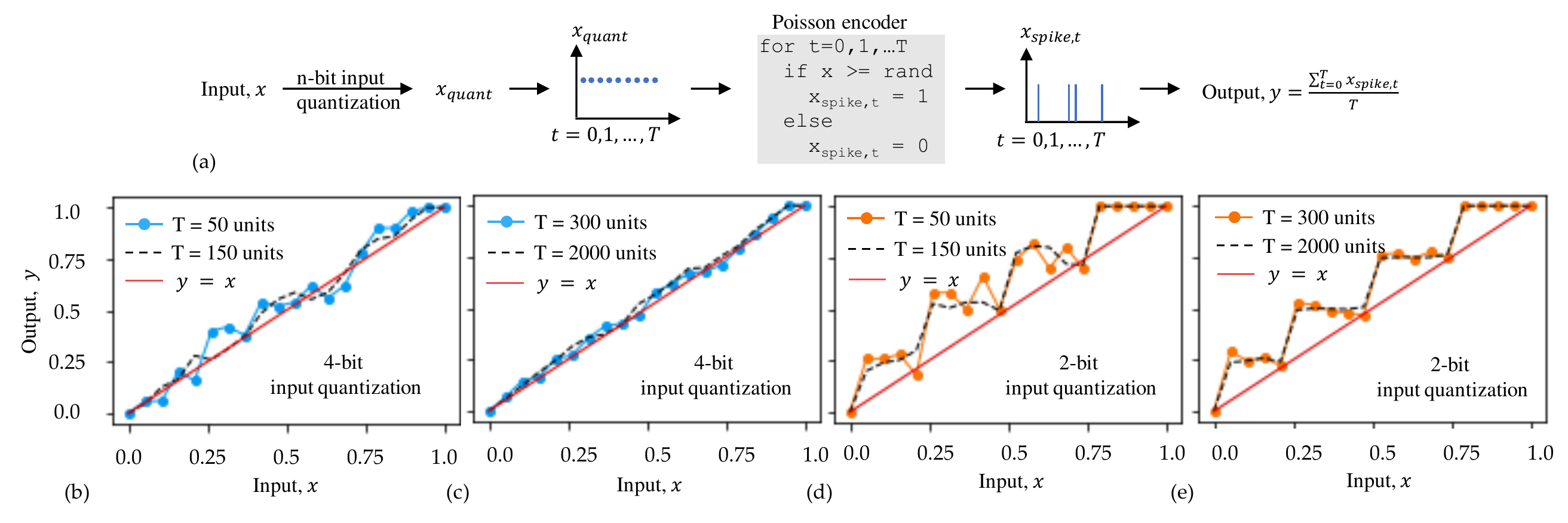}
\caption{The input-output characteristics of Poisson encoder to demonstrate the effect of the total number of timseteps $T$ used to generate the spike train with different levels of pre-quantization of the analog input. When T is in the low value (between 50 to 150) regime (subplots (b) and (d)), the amount of quantization significantly changes for varying the number of timesteps ({\it{solid dotted and dashed lines}}). But in the high value regime of $T$ (plots (c) and (e)), {\it{solid dotted}} and {\it{dashed}} lines almost coincide due to the averaging effect. The flow of data from output $y$ to input $x$ is showed in the schematic in (a)}
\label{fig:poisson_fig1}
\end{figure}
The reason behind the invariance of SNN-conv towards the number of timesteps is explained in Fig.~\ref{fig:poisson_fig1}. We plot the input-output characteristics of the Poisson-encoder for 4 cases: (b) 4-bit input quantization with smaller number of timesteps (50 and 150), (c) 4-bit quantization, larger number of timesteps (300 and 2000) and their 2-bit counterparts in (d) and (e), respectively. It is evident from (c) and (e) that larger number of timesteps introduces more of an averaging effect, than quantization, and hence, varying the number of timesteps has negligible effect on the transfer plots ({\it solid dotted} and {\it dashed} lines coincide), which is not true for (b) and (d). Due to this averaging effect, Poisson output $y$ for SNN-conv tends to follow the trajectory of $x_{quant}$ (quantized ANN input), leading to comparable adversarial accuracy to the corresponding ANN over the entire range of timesteps in Fig.~\ref{fig:poisson_fig2}(b). Note that, in these plots input $x$ refers to the analog input signal, whereas output $y$ is the timed average of the spike train (as showed in the schematic in Fig.~\ref{fig:poisson_fig1}(a)). 
\subsubsection{Effect of LIF (or IF) neurons and the leak factor}
\label{sec:neuron}
\begin{figure}[t]
\centering
\includegraphics[height=4.5cm]{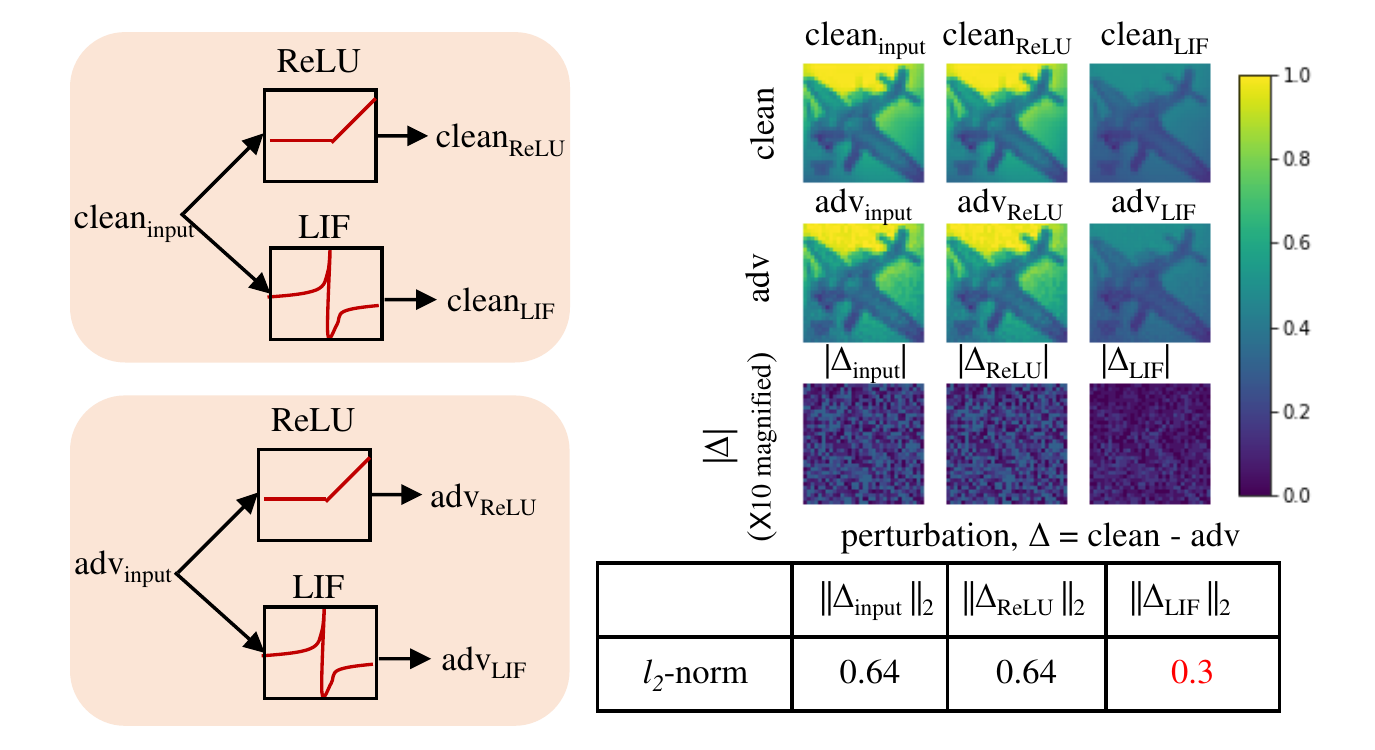}
\caption{The input and output of ReLU and LIF neurons for each of clean and adversarial image. Column 1 shows clean image, adversarial image and the absolute value of the adversarial perturbation before passing through the neurons. Column 2 and 3 depict the corresponding images after passing through a ReLU and an LIF, respectively. The bottom table contains the $l_2\mbox{-}norm$ of the perturbation at the input, ReLU output and LIF output}
\label{fig:relu_vs_lif}
\end{figure}
\begin{figure}[t!]
\centering
\includegraphics[height=4.5cm]{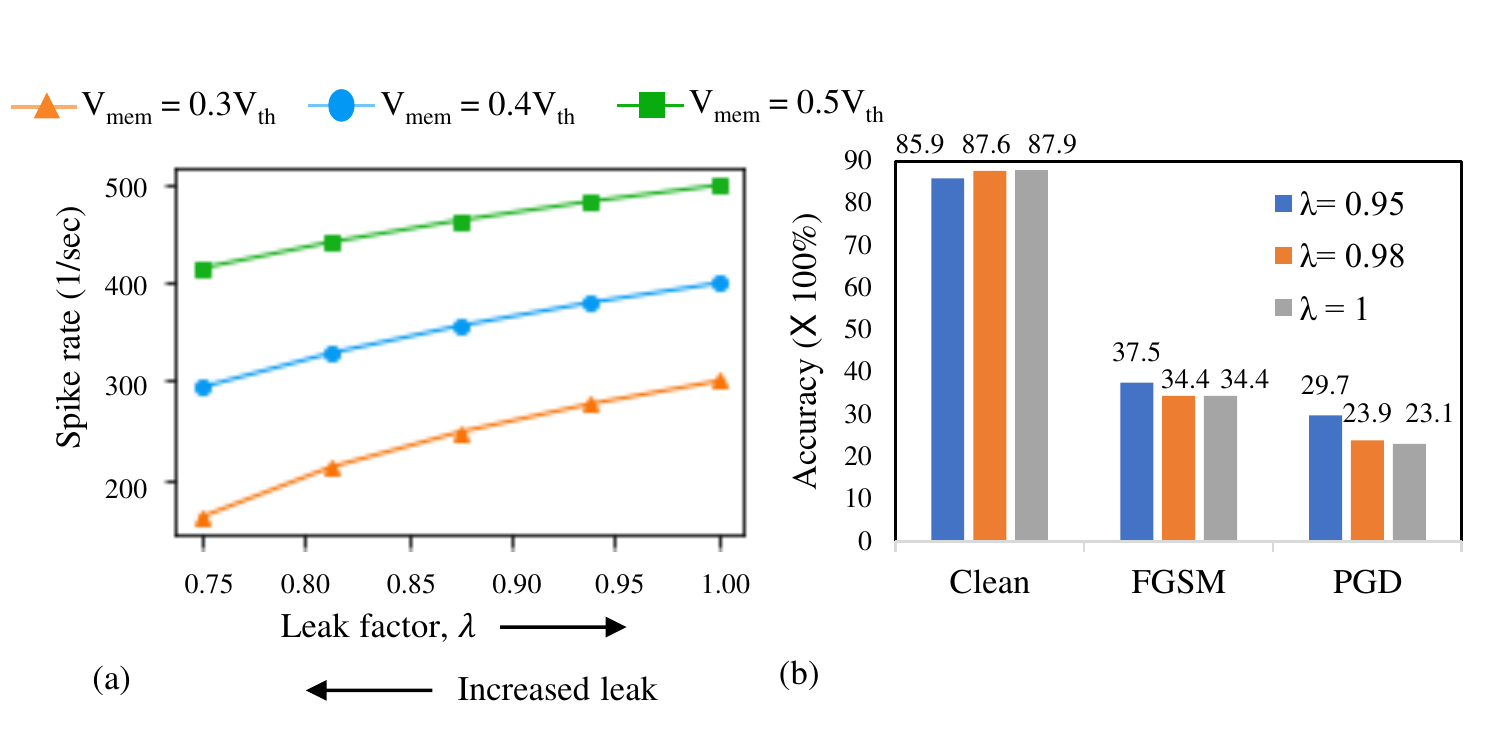}
\caption{(a) Spike rate versus leak factor $\lambda$ for different values of $\frac{V_{mem}}{V_{th}}$. Smaller value of $\lambda$ corresponds to more leak. (b) A bar plot showing the comparison of clean, FGSM and PGD($\epsilon = 8/255$) accuracy for a VGG5 SNN-BP network trained on CIFAR10 for different values of $\lambda$. These are blackbox attacks crafted from a VGG5 ANN model}
\label{fig:leak}
\end{figure}

Another major contributing factor towards SNN robustness is their highly nonlinear neuron activations (Integrate-Fire or Leaky-Integrate-Fire), whereas ANNs use mostly piecewise linear activations like ReLU. In order to explain the effect of this nonlinearity, we perform a proof of concept experiment. We feed a clean and corresponding adversarial input to a ReLU and an LIF neuron ($\lambda = 0.99$ in Eq.~\ref{eq:vt_lif}). Both of the inputs are $32 \times 32$ images with pixel intensity normalized to $[0,1]$. Row 1, 2 and 3 in Fig.~\ref{fig:relu_vs_lif} present the clean image, corresponding adversarial image and their absolute difference (amount of perturbation), respectively. 
Note, the outputs of the LIF neurons are binary at each timestep, hence, we take an average over the entire time-window to obtain corresponding pixel intensity. ReLU passes both clean and adversarial inputs without any transformation, hence the $l_2$-norm of the perturbation is same at the input and ReLU output (bottom table of the figure). However, the non-linear transformation in LIF reduces the perturbation of 0.6 at input layer to 0.3 at its output. Basically, the output images of LIF ({\it{column 3}}) neurons is a low pixel version of the input images, due to the translation of continuous analog values into a binary spike representation. This behavior helps diminish the propagation of adversarial perturbation through the network. IF neurons also demonstrate this non-linear transformation. However, the quantization effect is minimized due to their operation over longer time-window (as explained in the previous section).
\begin{wrapfigure}{r}{0.48\textwidth}
  \begin{center}
    \includegraphics[width=0.48\textwidth]{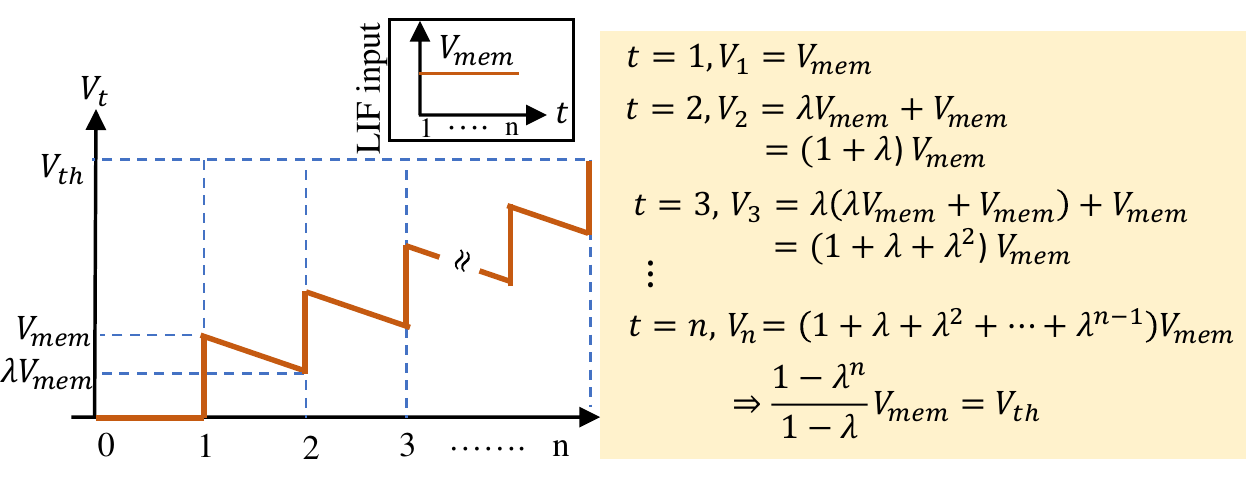}
  \end{center}
  \caption{Output of an LIF neuron for constant input voltage.}
  \label{fig:vmem}
\end{wrapfigure}
Unlike SNN-conv, SNN-BP networks can be trained with LIF neurons. The leak factor in an LIF neuron provides an extra knob to manipulate the adversarial robustness of these networks. In order to investigate the effect of leak on the amount of robustness, we develop a simple expression relating the leak factor with neuron spike rate in an LIF neuron. In this case, the membrane potential $V_t$ at timestep $t$ is updated  as $V_t = \lambda V_{t-1} + V_{input,t}$ given the membrane potential has not reached threshold yet, and hence, reset signal = 0. Here, $\lambda$ ($<1$) is the leak factor and $V_{input,t}$ is the input to the neuron at timestep $t$. Let us consider the scenario, where a constant voltage $V_{mem}$ is fed into the neuron at every timestep and the membrane potential reaches the threshold voltage $V_{th}$ after $n$ timesteps. As explained in Fig. \ref{fig:vmem}, membrane potential follows a geometric progression with time. After replacing $\frac{V_{th}}{V_{mem}}$ with a constant $r$, we obtain the following relation between the rate of spike ($1/n$) and leak factor ($\lambda$):
\begin{equation}
    \textrm{Spike rate,}\quad \frac{1}{n} = \frac{log\lambda}{log[r\lambda-(r-1)]}, \lambda<1
    \label{eq:spike_rate}
\end{equation}
In Fig.~\ref{fig:leak}(a), we plot the spike rate as a function of leak factor $\lambda$ for different values of $V_{mem}$ according to Eq.~\ref{eq:spike_rate}, where $\lambda$ is varied from 0.9999 to 0.75.
In every case, spike rate decreases ({\it{i. e.}} sparsity and hence, robustness increases) with increased amount of leak (smaller $\lambda$). The plot in Fig.~\ref{fig:leak}(b) justifies this idea where we show the adversarial accuracy of an SNN-BP network (VGG5 with CIFAR10) trained with different values of leak. For both FGSM and PGD attacks, adversarial accuracy increases by $3\sim6\%$ as $\lambda$ is decreased to 0.95. Note, the clean accuracy of the trained SNNs with different leak factors lies within a range of $\sim2\%$. In addition to sparsity, leak makes the membrane potential (in turn, the output spike rate) dependent on the temporal information of the incoming spike train \cite{leak_robustness}. Therefore, for a given input, while the IF neuron produces a deterministic spike pattern, the input-output spike mapping is non-deterministic in an LIF neuron. This effect gets enhanced with increased leak. We assume that this phenomenon is also responsible to some extent for the increased robustness of backpropagated SNNs with increased leak. It is worth mentioning here that Eq.~\ref{eq:spike_rate} holds when input to the neuron remains unchanged with the leak factor. In our experiments, we train SNN-BP with different values of $\lambda$ starting from the same initialized ANN-SNN converted network. Hence, the parameters of SNN-BP trained with different leak factors do not vary much from one another. Therefore, the assumption in the equation applies to our results.
\subsection{ANN-Crafted vs SNN-Crafted Attack}
\label{sec:annVSsnn_attack}
Lastly, we propose an attack-crafting technique from SNN with the aid of the surrogate gradient calculation. The details of the method is explained in Sec.~\ref{sec:adv_method}. Table~\ref{table:annVSsnn_attack} summarizes a comparison between ANN-crafted and SNN-crafted (our proposed technique) attacks. Note, these are blackbox attacks, {\it{i. e.}}, we train two separate and independently initialized models for each of the 3 networks (ANN, SNN-conv, SNN-BP). One of them is used as the source ({\it{marked as ANN-I, SNN-conv-I etc.}}) and the other ones as the target ({\it{marked as ANN-II, SNN-conv-II etc.}}). It is clear that SNN-BP adversarial accuracy (last row) is the highest for both SNN-crafted and ANN-crafted inputs. Moreover, let us analyze row 1 of Table~\ref{table:annVSsnn_attack} for FGSM attack. When ANN-II is attacked by ANN-I, FGSM accuracy is $18.9\%$, whereas, if attacked by an SNN-conv-I (or SNN-BP-I), the accuracy is $32.7\%$ (or $31.3\%$). Hence, these results suggest that ANN-crafted attacks are stronger than the corresponding SNN counterparts.
\newcolumntype{P}[1]{>{\centering\arraybackslash}p{#1}}
\begin{table}[t]
\begin{center}
\caption{A comparison of the blackbox adversarial accuracy for ANN-crafted {\it{versus}} SNN-crafted attacks. ANN-I and ANN-II are two separately trained VGG5 networks with different initializations. The same is true for SNN-conv and SNN-BP.  }
\label{table:annVSsnn_attack}
\begin{tabular}{|l||P{10mm}|P{10mm}|P{10mm}||P{10mm}|p{10mm}|P{10mm}|}
\hline\hline
&\multicolumn{3}{c}{FGSM} & \multicolumn{3}{c}{PGD}\\
\hline\hline
\backslashbox{Target}{Source} 
&{ANN-I} &{SNN-conv-I} &{SNN-BP-I}&{ANN-I} &{SNN-conv-I} &{SNN-BP-I} \\
\hline\hline
ANN-II &18.9\% &32.7\% &31.3\% &4.7\% &31.7\% & 13.8\%\\
SNN-conv-II &19.2\% &33.0\% &31.4\% &11.6\% &32.4\% & 14.3\%\\
SNN-BP-II &21.5\%&38.8\% &32.9\% &9.7\% &43.6\% &17.0\%\\
\hline
\end{tabular}
\end{center}
\end{table}

\section{Conclusions}
The current defense mechanisms in ANNs are incapable of preventing a range of adversarial attacks. In this work, we show that SNNs are inherently resilient to gradient-based adversarial attacks due to the discrete nature of input encoding and non-linear activation functions of LIF (or IF) neurons. The resiliency can be further improved by reducing the number of timesteps in the input-spike generation process and increasing the amount of leak of the LIF neurons. SNNs trained using ANN-SNN conversion technique (with IF neurons) require larger number of timesteps for inference than the corresponding SNNs trained with spike-based backpropagation (with LIF neurons). Hence, the latter technique leads to more robust SNNs. Our conclusion is validated only for gradient-based attacks on deep VGG and ResNet networks with CIFAR datasets. Future analysis on more diverse attack methods and architectures is necessary. We also propose a method to generate gradient-based attacks from SNNs by using the surrogate gradients.

\section*{Acknowledgement}
The work was supported in part by, Center for Brain-inspired Computing (C-BRIC), a DARPA sponsored JUMP center, Semiconductor Research Corporation, National Science Foundation, Intel Corporation, the DoD Vannevar Bush Fellowship and U.S. Army Research Laboratory.

%
%
\bibliographystyle{splncs04}
\bibliography{eccv2020}

\section*{Supplementary}
\subsection*{SNN-crafted adversarial example}
In this section, we develop a technique to produce adversarial examples from a Spiking Neural Network (SNN) utilizing the Fast Gradient Sign Method (FGSM) or its iterative variants.\\
Suppose the source model is referred as $M$, the loss function of the model for input $x$ and true labels $y_{true}$, is given by $J(x,y_{true})$. According to FGSM, the adversarial example is computed as,
\begin{equation}
    x_{adv} = x + \epsilon \times sign\big(\nabla_xJ\left(x,y_{true}\right)\big)
\end{equation}
$\epsilon$ denotes the strength of the perturbation. FGSM adversarial examples require computing the gradient of loss with respect to input ($\nabla_xJ(x,y_{true})$) obtained through backpropagation. Hence, it is essential to have a differentiable network from output all the way to input. However, SNNs have two sources of discontinuity in the gradient: (1) discontinuous gradients of the Leaky-Integrate-Fire (or Integrate-Fire) neurons in the hidden layers, (2) discretization at the input layer (Poisson encoding). We circumvent these two obstactles by the following approximations:
\begin{enumerate}
    \item The transfer function of LIF (or IF) neuron is a step function, the derivative being discontinuous at the threshold point. We approximate the gradients of the neurons by some pseudo-derivatives like linear or exponential functions. This approach, known as surrogate-gradient technique, is applied to all the hidden layers, thereby making the network differentiable from output to the $1st$ hidden layer.
    \item At the input layer, continuous-valued analog input is discretized into binary ($0$ or $1$) spike train. The analog input can be approximated by the average of this spike train over the entire time-window, referred as the rate input, $\mathbf{X_{rate}}$. We assume that $\mathbf{X_{rate}}$, which is a continuous-valued signal, is the input to the $1st$ convolutional layer.
\end{enumerate}
These two approximations make the SNN completely differentiable from output to input.
Let us consider an SNN with analog input $\mathbf{X}$ and input spike-train $\{\mathbf{X_1},\mathbf{X_2},\cdots \mathbf{X_T}\}$, where $T$ is the total number of timesteps. Each of $\mathbf{X_1}, \mathbf{X_2}, \cdots, \mathbf{X_T}$ has the same dimension as the analog input $\mathbf{X}$. $\mathbf{X_{rate}}$ describes the timed-average of the spike-train. The output of the $1st$ convolutional layer (weight matrix $\mathbf{W_{conv1}}$) passes through an LIF (or IF) neuron with membrane potential represented as $\mathbf{X_{conv1}}$ after $T$ timesteps. We assume that
\begin{equation}
    \mathbf{X} \approx \mathbf{X_{rate}}=\frac{\sum_{i=1}^{i=T}\mathbf{X_i}}{T}
\end{equation}
and
\begin{equation}
    \mathbf{X_{conv1}} \approx \mathbf{X_{rate}}\circledast \mathbf{W_{conv1}}
    \label{eq:conv_assumption}
\end{equation}
(The validity of the assumption in Eq.~\ref{eq:conv_assumption} is explained in Sec.~\ref{sec:validity}.) Let us assume the loss function of the network is represented by $J$. Then the gradient of loss with respect to input is described using the chain rule of derivative,
\begin{align}
    \frac{\partial J}{\partial \mathbf{X}} \approx \frac{\partial J}{\partial \mathbf{X_{rate}}} &= \frac{\partial J}{\partial \mathbf{X_{conv1}}} \times \frac{\partial \mathbf{X_{conv1}}}{\partial \mathbf{X_{rate}}} \\
    &= \frac{\partial J}{\partial \mathbf{X_{conv1}}} \circledast  \mathbf{W^{180rotated}_{conv1}}
    \label{eq:conv}
\end{align}
where $\mathbf{W^{180rotated}_{conv1}}$ is the $\mathbf{W_{conv1}}$ matrix rotated by \ang{180}. The proof of Eq.~\ref{eq:conv} is provided in the next subsection.
Hence, FGSM adversarial perturbation is computed as,
\begin{equation}
    \epsilon \times sign(\frac{\partial J}{\partial \mathbf{X}}) \approx \epsilon \times  sign(\frac{\partial J}{\partial \mathbf{X_{rate}}}) = \epsilon \times sign(\frac{\partial J}{\partial \mathbf{X_{conv1}}}\circledast \mathbf{W^{180rotated}_{conv1}})
    \label{eq:final}
\end{equation}
$\frac{\partial J}{\partial \mathbf{X_{conv1}}}$ is obatined by backpropagating the loss through the network with the help of the surrogate gradient technique. Then Eq.~\ref{eq:final} is used to obtain the adversarial perturbation.\\

\subsection*{Derivation of $\frac{\partial J}{\partial X}$ when $Y=X\circledast W$}
\label{sec:derivation}
Suppose the input image is represented as $\mathbf{X}$, $\mathbf{W}$ denotes the weight matrix of the convolutional layer and $\mathbf{Y}$ is the output of the convolution operation.
\begin{equation}
    \mathbf{Y} = \mathbf{X}\circledast\mathbf{W}
\end{equation}
Assume $\mathbf{X}$ is a $32\times32$ matrix and the kernel size $\mathbf{W}$ for convolution is $3\times3$. Convolution is performed with a padding of 1 and stride of 1. Then output $\mathbf{Y}$ will maintain a size of $32\times32$ $(\textrm{input dimension}+(2\times\textrm{padding})-\textrm{kernel size}+1)$.
\begin{equation*}
    \mathbf{X}=
    \begin{pmatrix}
    X_{1,1} & X_{1,2} & \cdots & X_{1,32} \\
    X_{2,1} & X_{2,2} & \cdots & X_{2,32} \\
    \vdots & \vdots & \ddots & \vdots \\
    X_{32,1} & X_{32,2} & \cdots & X_{32,32} \\
    \end{pmatrix}
    , \mathbf{W}=
    \begin{pmatrix}
    W_{1,1} & W_{1,2} & W_{1,3} \\
    W_{2,1} & W_{2,2} & W_{2,3} \\
    W_{3,1} & W_{3,2} & W_{3,3} \\
    \end{pmatrix}
\end{equation*}
\begin{equation}
    \mathbf{Y}=
    \begin{pmatrix}
    0 & 0 & 0 & \cdots & 0 & 0\\
    0 & X_{1,1} & X_{1,2} & \cdots & X_{1,32} & 0\\
    0 & X_{2,1} & X_{2,2} & \cdots & X_{2,32} & 0 \\
    \vdots & \vdots & \vdots & \ddots & \vdots & \vdots\\
    0 & X_{32,1} & X_{32,2} & \cdots & X_{32,32} & 0 \\
    0 & 0 & 0 & \cdots & 0 & 0\\
    \end{pmatrix}
    \circledast
    \begin{pmatrix}
    W_{1,1} & W_{1,2} & W_{1,3} \\
    W_{2,1} & W_{2,2} & W_{2,3} \\
    W_{3,1} & W_{3,2} & W_{3,3} \\
    \end{pmatrix}
\end{equation}
\begin{align*}
    Y_{1,1}&= W_{2,2}X_{1,1}+W_{2,3}X_{1,2}+W_{3,2}X_{2,1}+W_{3,3}X_{2,2}\\
    Y_{1,2} &= W_{2,1}X_{1,1}+W_{2,2}X_{1,2}+W_{2,3}X_{1,3}+W_{3,1}X_{2,1}+W_{3,2}X_{2,2}+W_{3,3}X_{2,3}\\
    \vdots \\
    Y_{32,32}&= W_{1,1}X_{31,31}+W_{1,2}X_{31,32}+W_{2,1}X_{32,31}+W_{2,2}X_{32,32}
\end{align*}
Let us assume the loss function is represented by $J$. The gradient of $J$ with respect to $\mathbf{X}$ is also a $32\times32$ matrix, each element described by the chain rule of derivative:
\begin{align}
    \frac{\partial J}{\partial X_{1,1}}&= \frac{\partial J}{\partial Y_{1,1}}\frac{\partial Y_{1,1}}{\partial X_{1,1}} +  \frac{\partial J}{\partial Y_{1,2}}\frac{\partial Y_{1,2}}{\partial X_{1,1}} + \hdots +  \frac{\partial J}{\partial Y_{32,32}}\frac{\partial Y_{32,32}}{\partial X_{1,1}} \\
    &=\frac{\partial J}{\partial Y_{1,1}}W_{2,2}+\frac{\partial J}{\partial Y_{1,2}}W_{2,1} +\frac{\partial J}{\partial Y_{2,1}}W_{1,2}+\frac{\partial J}{\partial Y_{2,2}}W_{1,1} \\
    \frac{\partial J}{\partial X_{1,2}}&= \frac{\partial J}{\partial Y_{1,1}}W_{2,3}+\frac{\partial J}{\partial Y_{1,2}}W_{2,2}+\frac{\partial J}{\partial Y_{1,3}}W_{2,1} +\frac{\partial J}{\partial Y_{2,1}}W_{1,3}+\frac{\partial J}{\partial Y_{2,2}}W_{1,2}+\frac{\partial J}{\partial Y_{2,3}}W_{1,1} \\
    &\vdots \\
    \frac{\partial J}{\partial X_{32,32}}&= \frac{\partial J}{\partial Y_{31,31}}W_{3,3}+\frac{\partial J}{\partial Y_{31,32}}W_{3,2}+\frac{\partial J}{\partial Y_{32,31}}W_{2,3} +\frac{\partial J}{\partial Y_{32,32}}W_{2,2}
\end{align}
\begin{align}
    &\frac{\partial J}{\partial \mathbf{X}}=
    \begin{pmatrix}
    0 & 0 & 0 & \cdots & 0 & 0\\
    0 & \frac{\partial J}{\partial Y_{1,1}} & \frac{\partial J}{\partial Y_{1,2}} & \cdots & \frac{\partial J}{\partial Y_{1,32}} & 0\\
    0 & \frac{\partial J}{\partial Y_{2,1}} & \frac{\partial J}{\partial Y_{2,2}} & \cdots & \frac{\partial J}{\partial Y_{2,32}} & 0 \\
    \vdots & \vdots & \vdots & \ddots & \vdots & \vdots\\
    0 & \frac{\partial J}{\partial Y_{32,1}} & \frac{\partial J}{\partial Y_{32,2}} & \cdots & \frac{\partial J}{\partial Y_{32,32}} & 0 \\
    0 & 0 & 0 & \cdots & 0 & 0\\
    \end{pmatrix}
    \circledast
    \begin{pmatrix}
    W_{3,3} & W_{3,2} & W_{3,1} \\
    W_{2,3} & W_{2,2} & W_{2,1} \\
    W_{1,3} & W_{1,2} & W_{1,1} \\
    \end{pmatrix}\\
    \therefore &\frac{\partial J}{\partial \mathbf{X}}=\frac{\partial J}{\partial \mathbf{Y}}\circledast \mathbf{W}^{180rotated},\quad\textrm{when}\quad \mathbf{Y}=\mathbf{X}\circledast\mathbf{W}
    \label{eq:delJ_delX}
\end{align}\\
\subsection*{Validity of the assumption $X_{conv1}\approx X_{rate}\circledast W_{conv1}$}
\label{sec:validity}
In an SNN, the membrane potential $\mathbf{X_{conv1}}$ at the final timestep $T$, after passing through  an IF neuron (leak factor $\lambda =1$), is computed as,
\begin{equation}
    \mathbf{X_{conv1}} = \sum_{i=1}^{i=T}\big(\mathbf{X_i}\circledast\mathbf{W_{conv1}}-\mathbf{V_{reset,i}}\big)
\end{equation}
$\mathbf{V_{reset,i}}$ is the reset voltage at $i\mbox{-}th$ timestep (in case of soft reset). $\mathbf{X_i}$ is the $i\mbox{-}th$ spike input. Suppose $V_{threshold}$ is the threshold voltage of the neuron and $\mathbf{X_{conv1,i}}$ is the membrane potential at $i\mbox{-}th$ timestep.
\begin{equation}
    V_{reset,i}=
    \begin{cases}
    0, & \text{if}\ X_{conv1,i}<V_{threshold} \\
    V_{threshold}, &\text{otherwise}
    \end{cases}
\end{equation}
Let us assume that the membrane potential reaches threshold $n$ times over the time-window $T$.
\begin{align}
    \mathbf{X_{conv1}} &= \sum_{i=1}^{i=T}\big(\mathbf{X_i}\circledast\mathbf{W_{conv1}}\big)-nV_{threshold} \\
    &=(\sum_{i=1}^{i=T}\mathbf{X_i})\circledast\mathbf{W_{conv1}}-nV_{threshold},\quad \textrm{distributive property of convolution}\\
    &=T\mathbf{X_{rate}}\circledast\mathbf{W_{conv1}}-nV_{threshold},\quad\textrm{T is the total number of timesteps}
    \label{eq:conv_eq}
\end{align}
Using the chain rule of derivative,
\begin{equation}
    \frac{\partial J}{\partial \mathbf{X_{rate}}} = \frac{\partial J}{\partial \mathbf{X_{conv1}}}\times \frac{\partial \mathbf{X_{conv1}}}{\partial \mathbf{X_{rate}}}
\end{equation}
Assuming the gradient of the second term in Eq.~\ref{eq:conv_eq} ({\it i.e.} $\frac{\partial (nV_{threshold})}{\partial \mathbf{X_{rate}}}$) to be negligibly small, we obtain the gradient of $J$ w.r.t. $\mathbf{X_{rate}}$ from Eq.~\ref{eq:delJ_delX} as,
\begin{equation}
    \frac{\partial J}{\partial \mathbf{X_{rate}}}\approx T(\frac{\partial J}{\partial \mathbf{X_{conv1}}}\circledast \mathbf{W^{180rotated}_{conv1}})
\end{equation}
Since $T$ is a positive constant value,
\begin{equation}
    sign(\frac{\partial J}{\partial \mathbf{X_{rate}}})=sign(\frac{\partial J}{\partial \mathbf{X_{conv1}}}\circledast \mathbf{W^{180rotated}_{conv1}})
\end{equation}
which complies with our solution at Eq.~\ref{eq:final}, obtained from the approximation $X_{conv1}\approx X_{rate}\circledast W_{conv1}$.
\end{document}